\documentclass[conference]{IEEEtran}
\IEEEoverridecommandlockouts
\usepackage{cite}
\usepackage{amsmath,amssymb,amsfonts}
\usepackage{algorithmic}
\usepackage{graphicx}
\usepackage{textcomp}
\usepackage{adjustbox}
\usepackage{bbm}
\usepackage{comment}
\usepackage{xcolor}
\usepackage{mathtools}

\def\BibTeX{{\rm B\kern-.05em{\sc i\kern-.025em b}\kern-.08em
    T\kern-.1667em\lower.7ex\hbox{E}\kern-.125emX}}
\usepackage{fancyhdr}
\begin{document}

\title{Domain-Invariant Feature Alignment Using Variational Inference For Partial Domain Adaptation}

\author{\IEEEauthorblockN{Sandipan Choudhuri, Suli Adeniye, Arunabha Sen, Hemanth Venkateswara}
\text{Arizona State University}\\
\{s.choudhuri, sadeniye, asen, hemanthv\}@asu.edu}

\maketitle

\begin{abstract}
The standard closed-set domain adaptation approaches seek to mitigate distribution discrepancies between two domains under the constraint of both sharing identical label sets. However, in realistic scenarios, finding an optimal source domain with identical label space is a challenging task. Partial domain adaptation alleviates this problem of procuring a labeled dataset with identical label space assumptions and addresses a more practical scenario where the source label set subsumes the target label set. This, however, presents a few additional obstacles during adaptation. Samples with categories private to the source domain thwart relevant knowledge transfer and degrade model performance. In this work, we try to address these issues by coupling variational information and adversarial learning with a pseudo-labeling technique to enforce class distribution alignment and minimize the transfer of superfluous information from the source samples. The experimental findings in numerous cross-domain classification tasks demonstrate that the proposed technique delivers superior and comparable accuracy to existing methods. 
\end{abstract}

\section{\textbf{Introduction}}

A broad spectrum of frameworks that address complex machine learning issues have demonstrated notable performance improvements, attributable to the deep neural networks \cite{liu2021review, wang2019development, dang2019deep, choudhuri2018object, guo2021survey}. For such models to be generalizable, large amounts of labeled data must be readily available for supervision. Procuring such heavily annotated data is challenging in some real-world scenarios when data gathering and subsequent annotation incur significant expenses. A domain adaptation $(da)$ strategy \cite{ganin2016domain} can reduce this annotation requirement by transferring relevant information from a large-scale dataset previously labeled and from a related domain.

\begin{figure}[htbp!]
    \centering
    \includegraphics[width = 0.85\linewidth, height = 6.5 cm]{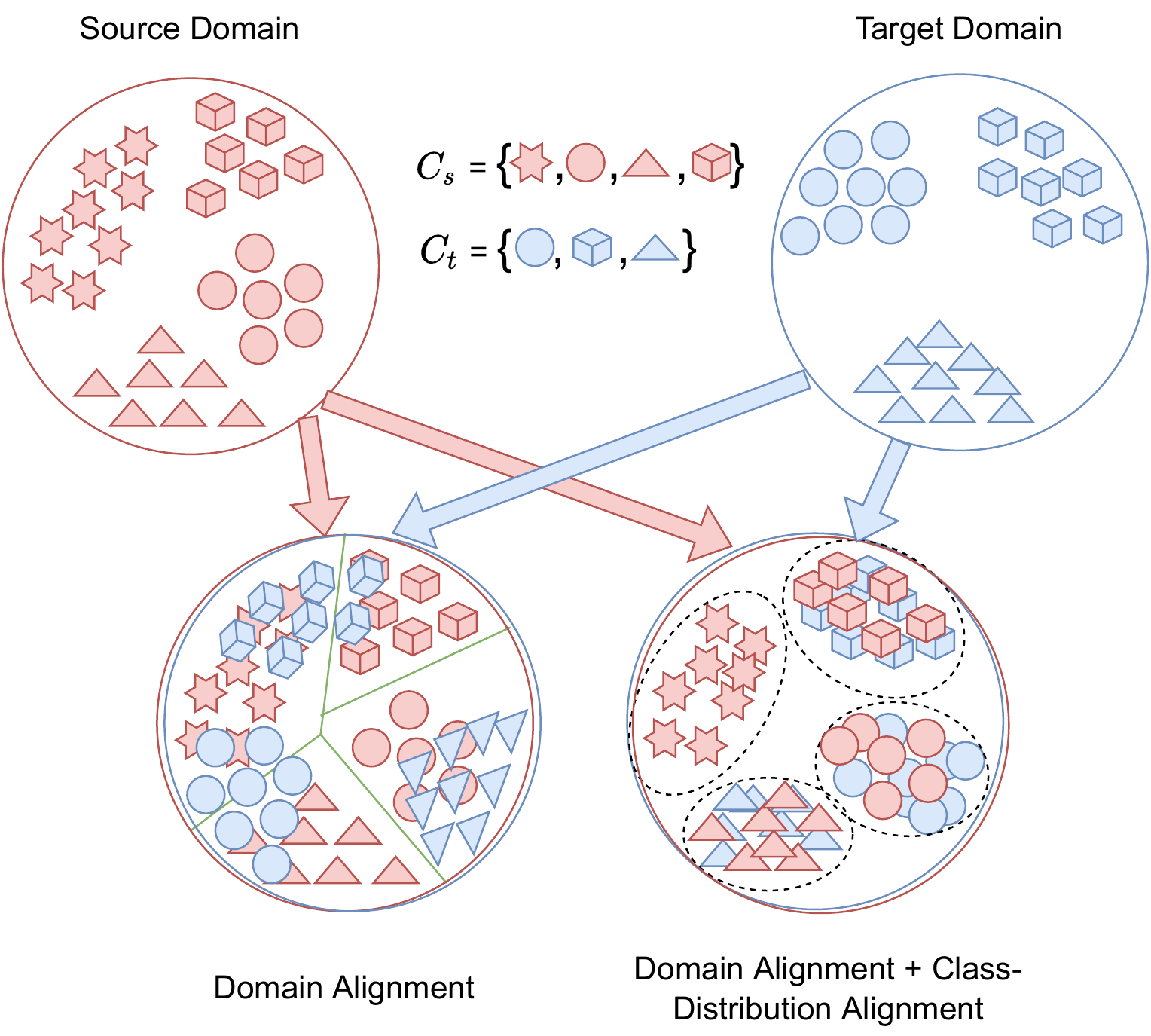}
    \caption{The figure above represents the latent space of the two domains. The source (represented in red) and target (represented in blue) domains contain data samples from four and three classes, respectively. The bottom left figure represents an instance where domain alignment is not a sufficient condition for improving classification accuracy. The bottom right figure represents a desirable instance for classification where elements from the same category are clustered to their respective distribution, and that from different categories are assigned to different distributions, regardless of their domains.}    
    \label{fig_1}
    \vspace{-5mm}
\end{figure}

The standard closed-set unsupervised domain adaptation $(uda)$ frameworks\cite{long2015learning,ganin2016domain}, which learn a classifier for the unlabeled target domain using a labeled source domain, have gained massive traction in the machine learning community. However, most existing works on $uda$ assume that the source and target domains have the same label set. Finding an optimal source domain with an identical label space is challenging in practical scenarios. A more feasible approach is to operate on a relatively small-scale target domain while accessing a large-scale source domain. Partial domain adaptation $(pda)$ \cite{cao2018partial,zhang2018importance,cao2018partial2,Choudhuri} addresses such a scenario, where the target label set is contained in the source label set. The following section discusses the current challenges in a $pda$ problem and our recommendations for mitigating them.



Prior works on \textit{pda} \cite{cao2018partial,zhang2018importance,cao2018partial2,cao2019learning,Choudhuri} have attempted to find shared latent representations of the source and target samples with class-discriminative properties. Among these, domain adversarial training is widely utilized for extracting domain-invariant latent features from the source and target samples, owing to its performance and extensibility. The process is performed with a feature extractor, a domain discriminator, and a label classifier. The latter two process the feature extractor output to predict domain and class labels. Attaining domain invariance, however, is a necessary and not a sufficient condition; ensuring improvement of target classification performance requires mitigation of the conditional distribution mismatch across two domains. Therefore, the latent space should be sufficiently ``well-organized'' and ``regular'' so that samples with the same class label are clustered to their respective distribution, while data with different class labels are assigned to distinct class distributions, regardless of their domains (check figure \ref{fig_1}). Furthermore, it is vital to ensure that the information captured in latent features of samples adheres to target data for exercising sufficient supervision when adapting to unlabeled target samples. In other words, two neighboring points in the latent space representing target data should not yield radically different class-specific contents. 

We have incorporated domain adversarial training to address these critical issues while enforcing explicit regularization of encoded sample data through variational information. The domain invariant features in the latent space are modeled as a mixture of Gaussian distributions, each representing the latent feature distribution of a predicted class. In addition, the model approximates a posterior feature distribution, where the latent features of a sample follow a Gaussian distribution. The model aims to align these posterior embeddings with the reference latent features during training. Enforcing this regularization assists the adapted model in minimizing inter-class entanglement and promotes class-wise distribution alignment in the latent space while capturing class-semantic information.

\begin{figure*}[htbp!]
    \centering
    \includegraphics[width = .9\linewidth]{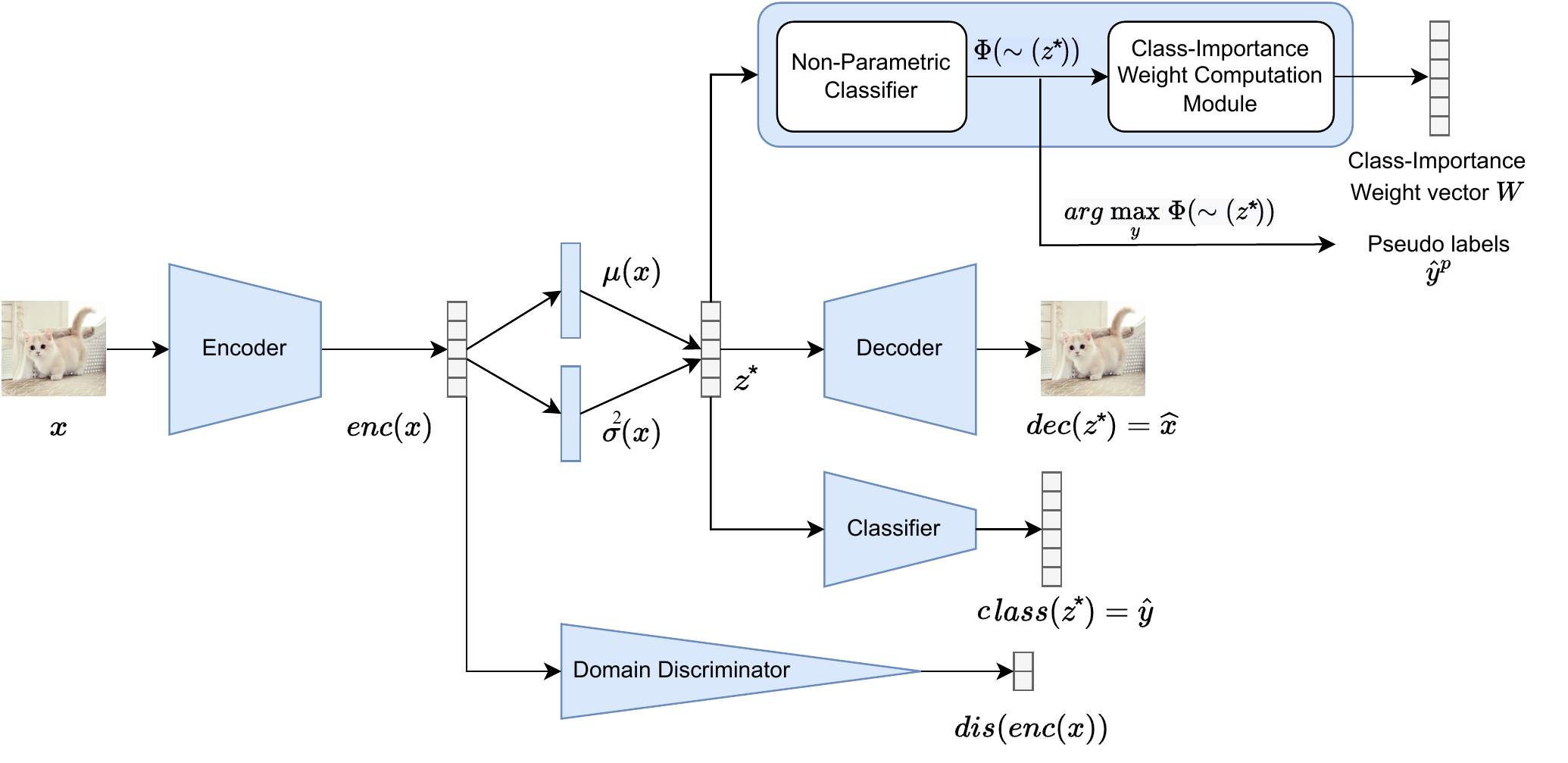}
    \caption{A schematic pipeline of the proposed network: Input sample $x$ is processed by an encoder $enc(x)$. The encoder output is passed through the domain discriminator, $dis(enc(x))$, which determines the domain membership of $x$. $enc(x)$ is also processed by networks $\mu(\cdot)$ and $\sigma^2(\cdot)$, to obtain feature means $\mu(enc(x))$ and feature variances $\sigma^2(enc(x))$, that   parameterize a Gaussian distribution. This distribution is sampled to obtain a latent representation $z^*$ for $x$. $z^*$ is subsequently passed through a decoder $dec(z^*)$ for data reconstruction $\hat{x}$, and a classifier for label prediction \big($\underset{y_s} {\mathrm{argmax}} \hspace{2mm} class(z^*) \leftarrow \hat{y_s}$\big). The encoder and classifier estimate the posterior distributions, while the decoder is utilized to infer the prior distributions. The non-trainable pipeline, consisting of a non-parametric classifier and a class-importance weight computation module, generates pseudo-labels for target data through a similarity function that quantifies a target sample's closeness to the representative centers of each source class. The class-weight importance estimation is performed through a voting strategy that involves the participation of a subset of target samples (ones with high-confidence predictions in the non-parametric classification).}    
    \label{fig_2}
    \vspace{-4mm}
\end{figure*}

Removing the constraint of identical label set assumptions between two domains introduces the risk of negative transfer (\textit{propagating of unwanted information from samples in classes private to the source domain}) into the model \cite{cao2018partial,zhang2018importance,cao2018partial2,cao2019learning}, and consequently thwarts classification performance. As the model is not initially privy to the knowledge of shared label-set between two domains in a \textit{pda} setup, it is essential to incorporate a mechanism into our network that estimates the common categories between two domains. Citing the necessity of eliminating negative transfer, we have designed a technique by quantifying the transferability of the source samples and regulating class-wise contribution to the learning of the classifier, domain discriminator, and feature decoder. This class-weighing scheme is further refined by filtering out confident task-relevant target samples for effective cross-domain alignment.

\section{\textbf{Related Work}}

Several studies \cite{hoffman2014lsda,oquab2014learning,yosinski2014transferable} in recent years have thoroughly explored the efficacy of deep neural networks for reducing domain discrepancy and effectively transferring relevant knowledge between domains for transfer learning tasks. A line of work \cite{zhang2018unsupervised} proposes a strategy for successfully aligning the distribution across domains and reducing domain discrepancy by applying high-order statistical features (primarily centered on \textit{maximum-mean discrepancy}). The authors of \cite{ganin2016domain, li2019joint} use adversarial learning to develop a mini-max game that extracts domain-invariant features by utilizing samples from common and private categories of the source dataset. They are, unfortunately, inefficient in a partial-domain adaptation environment and only effective in a limited, closed-set domain adaptation scenario.

By leveraging multiple adversarial networks to down-weight private source category samples, the Selective Adversarial Network (SAN) \cite{cao2018partial} handles partial-domain adaptation tasks and ensures efficient knowledge transfer. By expanding on this idea, the authors of \cite{cao2018partial2} provide a framework for class-importance weight estimation by combining target sample prediction scores. Similar ideas are put out by Zhang et al. \cite{zhang2018importance} in their work on Importance Weighted Adversarial Nets (IWAN), which makes use of an auxiliary domain discriminator to gauge how closely related a source sample is to the target domain. A soft indicator for distinguishing the common categories from the private source classes is proposed by the Example Transfer Network (ETN) \cite{cao2019learning}, which employs discriminative information to assess the transferability of source domain samples. 

Despite outperforming closed-set domain adaptation strategies, these models may have considerable limitations when determining the categories of private sources due to poor classification performance during the early training stages. With this work, we've attempted to address the limitations mentioned above.

\section{\textbf{Proposed Approach}}

\subsection{\textbf{Problem Definition}}

An \textit{Unsupervised Domain-Adaptation} $(uda)$ scenario assumes samples representing the source $s$ and target $t$ domains are drawn from different probability distributions\cite{Sugiyama}. As witnessed in a standard \textit{uda} environment, we are furnished with a source dataset $D_s=\{(x^i_s,y^i_s)\}_{i=1}^{n_s}$ of $n_s$ labeled points, sampled from a distribution $\mathbb{P}_s$, and an unlabeled target dataset $D_t=\{x^j_t\}_{j=1}^{n_t}$ of $n_t$ samples, drawn from distribution $\mathbb{P}_t$ \big($x^i_s,x^j_t \in \mathbb{R}^{d},\hspace{1mm} \mathbb{P}_s \neq \mathbb{P}_t$\big). Since target class label information is unavailable during adaptation, the \textit{Closed-Set} variant assumes that the samples in $D_s$ and $D_t$ are categorized into classes from known label-sets $C_s$ and $C_t$ respectively, where $C_s = C_t$. \textit{Partial domain adaptation} $(pda)$ generalizes this characterization and addresses a realistic scenario by alleviating the constraint of identical label space assumptions between the two domains (i.e., $C_t \subseteq C_s$).

With the objective of designing a classifier hypothesis $\mathcal{F}: x_t \rightarrow y_t$ that minimizes the target classification risk under a \textit{pda} setup, we aim at leveraging source domain supervision to capture class-semantic information, along with minimizing misalignment due to negative transfer from samples in the outlier label-space $C_s - C_t$  \big($\{(x_s,y_s)| (x_s,y_s) \in D_s \cap \hspace{1mm} y_s \in C_s - C_t$ \big\}\big).

\subsection{\textbf{Partial Domain Adaptation Model}}

With the objective outlined above, in this section, we present an overview of the proposed architecture. The learning process can be categorized into four major components, namely:

\begin{itemize}
    \item Attaining domain-invariance in the latent space.
    \item Establishing class-wise distribution alignment.
    \item Ensuring supervision of target samples through pseudo-label generation.
    \item Minimizing negative knowledge transfer from source samples in $C_s - C_t$ by regulating sample-wise contribution to classification, domain discrimination, and input reconstruction tasks.
\end{itemize}

The proposed model (fig. \ref{fig_2}) accepts an input sample $x$ from the source/target domain ($x \in \mathbb{R}^{d}$) and encodes it to a lower dimensional latent representation $enc(x)$ ($enc(x) \in \mathbb{R}^{d'}$, $d' < d$). The output of the encoder $enc(\cdot)$ is accepted by the domain discriminator, $dis(enc(x)) \in [0,1]$, which determines its domain membership. Concurrently, the encoder output is processed by networks $\mu(\cdot)$ and $\sigma^2(\cdot)$ to obtain the feature means and feature variances, respectively. $\mu(enc(x))$ and $\sigma(enc(x))$  parameterize a Gaussian distribution to obtain a latent feature sample $z^*$ of $x$ (the sampling process is conducted using the \textit{re-parameterization} trick). $z^*$ is subsequently passed through a decoder $dec(\cdot)$ and a classifier $class(\cdot)$ for data reconstruction ($\hat{x} \leftarrow dec(z^*)$) and label prediction ($\hat{y} \leftarrow \underset{y} {\mathrm{argmax}} \hspace{2mm} class(z^*)$), respectively. In addition, we utilize a pseudo-labeling strategy using non-parametric classifiers for supervision of target sample classification and computation of class-importance weights (required for reducing the effect of samples from outlier classes $C_s-C_t$). The following sections discuss the network architecture from the standpoint of mitigating the issues mentioned earlier. 

\subsubsection{\textbf{Domain-Invariant Feature Extraction with Adversarial Learning}}
\label{W_adv}

The adversarial approach for domain adaptation usually centers around matching the source and target feature distributions through a two-player minimax game. The idea has resulted in developing a series of DANNs (domain adversarial neural networks) \cite{Ganin}, which achieve high performance in a typical domain adaptation setup with shared label space across domains. In the proposed setup, the first player $dis(\cdot)$ is modeled as a domain discriminator and is trained to separate the $x_s$ from $x_t$. $enc(\cdot)$ poses as the second player trained to confuse the domain discriminator simultaneously by generating domain-invariant features. The encoder weights are learned by maximizing the loss of $dis(\cdot)$, whereas the discriminator weights are learned by reducing the loss of $dis(\cdot)$ to extract domain-transferable features $z$. 

The overall objective of the Domain Adversarial Neural Network is realized by minimizing the following term:
\begin{multline}
    L_{adv} = -\frac{1}{n_s}\sum\limits_{x_s \in D_s} w_{y_s}[ log \hspace{2mm}dis(enc(x_s))]\\ -\frac{1}{n_t}\sum\limits_{x_t \in D_t}[1 - log \hspace{2mm}dis(enc(x_t))] \text{,  } w_{y_s}\in W 
    \label{L_adv}
\end{multline}
With the objective of eliminating negative transfer, we down-weight the contributions of all outlier source samples from the source label space $C_s-C_t$. 
This is achieved by multiplying $w_{y_s}$ to the log value of domain discriminator output over the source domain data ($y_s$ is the ground truth label of source sample $x_s$, $w_{y_s}$ represents the corresponding class weight in the class-importance weight vector $W$). The detailed process of estimating $W$ is presented in section \ref{W_estimation}.

\subsubsection{\textbf{Latent Feature Alignment using Variational Information}}

As highlighted earlier, improving class conditional distribution alignment forms a salient task in \textit{pda} besides attaining domain invariance. The underlying classification objective is better realized if samples with the same class labels are mapped to the same reference distribution while samples with different class labels are assigned to different distributions. We propose to address this by regularizing the latent space. In this work, the latent features are modeled as a mixture of Gaussian distributions 

\begin{equation}
    p(z|y) = \mathcal{N}(z|\mu_{y}, \sigma^2_{y} I), \hspace{2mm} \forall y \in C_s 
\end{equation}
 In the equation above, $I$, $\mu_{y}$, and $\sigma^2_{y}$ signify the identity matrix, mean, and variance parameters, respectively. Here, each $p(z|y)$ represents a reference feature distribution (prior) for a predicted class $y$ (${\mathrm{argmax}} \hspace{2mm} class(\cdot)$).

The latent representations $z^*$ sampled from these distributions are subsequently processed for classification and data reconstruction to preserve class-discriminative and structural information in them. The reconstruction process is modelled with a Gaussian distribution $p(x|z^*) = \mathcal{N}(x|g(z^*),cI)$. The mean of this distribution is represented by the output of a deterministic function $g(\cdot)$ on $z^*$, while the covariance matrix is defined as $c$ times the identity matrix $I$ ($c$ is a positive constant). We approximate this distribution using a decoder neural network $dec(\cdot)$ where:
\begin{equation}
p(x|z^*) = \mathcal{N}(x|dec(z^*),I)
\end{equation}
With the assumptions of latent features as samples drawn from a mixture of Gaussian distributions, we aim to estimate the posterior distribution $p(y,z^*|x)$. Citing the potential of variational inference for learning latent representations \cite{Kingma}, we utilize it in our work to approximate $p(y, z^*|x)$ with $q(y,z^*|x)$. Assuming $x$ and $y$ are conditionally independent given $z^*$, we can expand the approximated distribution as $q(y,z^*|x) = q(y|z^*)q(z^*|x)$. The product terms in the $r.h.s$ signify classification and encoding, respectively. To learn smooth latent features, $q(z^*|x)$ is formulated as a sample-wise Gaussian distribution.
\begin{equation}
q(z^*|x) = \mathcal{N}(z^*|\mu(enc(x)), \sigma^2(enc(x))I)
\label{encoder}
\end{equation}

where $enc(.)$, $\mu(.)$, $\sigma^2(.)$ are neural networks. The classifier objective is realized by $q(y|z^*)$, as:
\begin{equation}
q(y|z^*) = class(z^*)
\end{equation}
Here, $class(.)$ represents the classifier neural network, followed by a $softmax$.
A pseudo-labeling strategy using a non-parameterized classifier is employed to obtain labels $\hat{y}_t$ for samples $x_t$ in $D_t$. A subset $D_t^{\mathcal{T}}\subseteq D_t$ of target samples with above-average confidence predictions are filtered out for training $class(\cdot)$, \textit{(illustrated further in section \ref{W_estimation})}. 
With the class information obtained for supervision,
the adapted predictions are matched to the class predictions for capturing class semantic information, using cross-entropy loss $CL(\cdot||\cdot)$.
\begin{multline}
    L_{class} = \frac{1}{n_s}  \sum_{\mathclap{\substack{(x_s,y_s) \in D_s\\z^*_s \sim q(z|x=x_s)}}} w_{y_s} CL(class(z^*_s), y_s)\\
    + \frac{1}{|D_t^{\mathcal{T}}|}  \sum_{\mathclap{\substack{(x_t,\hat{y}^p_t) \in D_t^{\mathcal{T}}\\z^*_t \sim q(z|x=x_t)}}} CL(class(z^*_t), \hat{y}^p_t)
\end{multline}

After establishing the reference (prior) and posterior distributions, we follow a variant of the distribution alignment strategy \cite{Yeh} inferred from maximizing the evidence lower bound and employ variational inference by aligning the encoded latent feature distributions $q(z^*|x)$ with the mixture of Gaussians $p(z^*|y)$. In addition, the adapted reconstructions are matched with the input sample using $l_2-norm$ to preserve the target information in the encoded representations. Using a similar strategy as utilized during adversarial domain alignment, we down-weight the contributions of all outlier source samples from the source label space using $W$ (presented in section \ref{W_estimation}). The combined objective encompassing the class distribution alignment ${L_{cda}}$ and data reconstruction $L_{recon}$, captured in $L_{var}$, is presented as: 

\begin{equation}
 L_{var} = L_{recon} + L_{cda},
 \label{l_var}
\end{equation}
\begin{multline}
\text{where, }L_{recon} = \frac{1}{n_s}  \sum_{\mathclap{\substack{(x_s,y_s) \in D_s\\z^*_s \sim q(z|x=x_s)}}} w_{y_s} || dec(z^*_s) - x_s|| \\
 +\frac{1}{n_t}  \sum_{\mathclap{\substack{x_t \in D_t\\z^*_t \sim q(z|x = x_t)}}} || dec(z^*_t) - x_t||, \text{ } w_{y_s} \in W
\end{multline}
\begin{equation}
\text{and, }L_{cda} = \sum_{\mathclap{\substack{y \in C_s\\z^* \sim q(z|x)\\x \in D_s \cup D_t^{\mathcal{T}}}}} \bigg[q(y \vert z^*) log\bigg( \frac{q(z^* \vert x)}{p(z^* \vert y)}\bigg)\bigg]
\end{equation}

\subsubsection{\textbf{Target Supervision and Estimation of Class-Importance Weights through Pseudo-Labels}}
\label{W_estimation}

The classification performance in a $pda$ setup is contingent upon the model's capability to limit negative transfer from the private category samples in $s$. The extraneous information contained in these samples might confuse the classifier, resulting in an increase in classification error. Therefore, a filtration mechanism is necessary to limit their contribution to the learning process. Most of the existing solutions to the $pda$ problem \cite{cao2018partial,zhang2018importance,cao2018partial2,cao2019learning} attempt to address the negative transfer issue by re-weighting samples with their predicted classification probabilities or by performing a class-wise aggregation of all the target samples for estimating the shared classes. These are, however, not satisfactory strategies and may induce severe classification errors, thereby misleading the optimization process. The effect is especially drastic during the initial stages of training when the classifier is shallow-trained and does not generate predictions with high confidence. 

This work utilizes a subset of target samples in the class-importance weight computation process by leveraging high-confidence target predictions. Inspired by the domain adaptation approaches presented in \cite{Mingsheng,Long,choudhuri2022coupling}, we enable target domain supervision by incorporating pseudo-labels generated from a non-parametric classifier. The adopted pseudo-labeling strategy is as follows:

\begin{itemize}
    \item \textbf{Step 1:} For each input sample $x_s$ in  $D_s$ and $x_t$ in $D_t$, we obtain their encoded latent representations $z_s^{*}$, $z_t^{*}$ respectively, using eq. \ref{encoder}.

    \item \textbf{Step 2:} Using the encoded representation $z_s^{*}$ of a source sample $x_s$ and it's corresponding category information $y_s$, we compute the cluster centers $\{\mu^{c_s}_s\}$ $\forall c_s \in C_s$, where:
    \begin{equation}
    \begin{gathered}
        \mu^{c_s}_s = \frac{1}{n^{c_s}_s}\sum_{x^{c_s}_s \in D^{c_s}_s} z^{*{c_s}}_s\\
        \text{where, } z^{*{c_s}}_s \sim \mathcal{N}(\mu(enc(x^{c_s}_s)), \sigma^2(enc(x^{c_s}_s))I),\\
        x^{c_s}_s \in D^{c_s}_s = \{x_s | (x_s,y_s) \in D_s \cap \hspace{1mm} y_s=c_s\},\\
        \text{and } n^{c_s}_s = |D^{c_s}_s|
    \end{gathered}
    \end{equation}
    
    \item \textbf{Step 3:} A similarity function $sim({\cdot})$ (returns a vector of size $|C_s|$) is computed for each $x_t$, that quantifies its closeness to the representative centers of each source class and is represented as:
    
    \begin{equation}
        sim(x_t) = \bigg[\frac{2-JS(z_t^{*}||\mu^{c_s}_s)}{2}\bigg]_{c_s \in C_s}
    \end{equation}
    We formulate a similarity measure using the Jensen-Shannon divergence metric \big($JS(.)$\big) to measure the closeness between the latent-target vector and cluster centers of the latent-source representations. The final value for each entry is normalized in the range [0,1], with a higher value signifying greater similarity.
    \item \textbf{Step 4:} Probability predictions are assigned for $x_t$ by computing \textit{softmax}, ${\Phi}(.)$, over similarity values in $sim_{x_t}$. i.e.:
    \begin{equation}
        \hat{p}_t = \hspace{2mm} {\Phi}(sim(x_t))
        \label{softmax_prob}
    \end{equation}
    \begin{equation}
        \hat{y}^p_t = \underset{c_s} {\mathrm{argmax}} \hspace{2mm} \big[{\Phi}(sim(x_t))\big]
        \label{pseudo_labels}
    \end{equation}
    
    In eq. \ref{pseudo_labels}, $\hat{y}^p_t$ represents the predicted pseudo-label for $x_t$, essential for classifier training on target samples. $\hat{p}_t$, in eq. \ref{softmax_prob}, is a vector of \textit{softmax} probability values, with $k^{th}$ entry representing the probability of $x_t$ belonging to class $k \in C_s$.  
\end{itemize}

\begin{table*}[htbp!]
\begin{adjustbox}{width=1\textwidth}

 \begin{tabular}{c | c c c c c c c c c c c c | c} 
 \hline
 Method & Ar $\rightarrow$ Cl & Ar $\rightarrow$ Pr & Ar $\rightarrow$ Rw & Cl $\rightarrow$ Ar & Cl $\rightarrow$ Pr & Cl $\rightarrow$ Rw & Pr $\rightarrow$ Ar & Pr $\rightarrow$ Cl & Pr $\rightarrow$ Rw & Rw $\rightarrow$ Ar & Rw $\rightarrow$ Cl & Rw $\rightarrow$ Pr & Avg. \\ 
 \hline
Resnet-50\cite{he2016deep} & 46.33 & 67.51 & 75.87 & 59.14 & 59.94 & 62.73 & 58.22 & 41.79 & 74.88 & 67.40 & 48.18 & 74.17 & 61.35\\
DANN\cite{ganin2016domain} & 43.76 & 67.90 & 77.47 & 63.73 & 58.99 & 67.59 & 56.84 & 37.07 & 76.37 & 69.15 & 44.30 & 77.48 & 61.72\\
ADDA\cite{tzeng2017adversarial} & 45.23 & 68.79 & 79.21 & 64.56 & 60.01 & 68.29 & 57.56 & 38.89 & 77.45 & 70.28 & 45.23 & 78.32 & 62.82\\
PADA\cite{cao2018partial2} & 51.95 & 67.00 & 78.74 & 52.16 & 53.78 & 59.03 & 52.61 & 43.22 & 78.79 & 73.73 & 56.60 & 77.09 & 62.06\\
SSPDA\cite{cao2019learning} & 52.02 & 63.64 & 77.95 & 65.66 & 59.31 & 73.48 & 70.49 & 51.54 & \textbf{84.89} & 76.25 & 60.74 & \textbf{80.86} & 68.07\\
RTN\cite{long2016unsupervised} & 49.31 & 57.70 & 80.07 & 63.54 & 63.47 & 73.38 & 65.11 & 41.73 & 75.32 & 63.18 & 43.57 & 80.50 & 63.07\\
IWAN\cite{zhang2018importance} & 53.94 & 54.45 & 78.12 & 61.31 & 47.95 & 63.32 & 54.17 & 52.02 & 81.28 & 76.46 & 56.75 & 82.90 & 63.56\\
SAN\cite{cao2018partial} & 44.42 & 68.68 & 74.60 & \textbf{67.49} & \textbf{64.99} & \textbf{77.80} & 59.78 & 44.72 & 80.07 & 72.18 & 50.21 & 78.66 & 65.30\\

 \hline
 
Proposed model & \textbf{54.18} & \textbf{69.22} & \textbf{81.44} & 65.91 & 64.73 & 73.81 & \textbf{71.26} & \textbf{52.31} & 83.93 & \textbf{76.48} & \textbf{60.92} & 81.04 & \textbf{69.60} \\
w/o ast & 52.87 & 68.78 & 79.16 & 63.92 & 63.88 & 70.01 & 69.17 & 50.21 & 79.86 & 74.38 & 58.16 & 79.62 & 67.50 \\
w/o adv & 48.53 & 69.74 & 77.38 & 60.97 & 62.39 & 64.92 & 62.13 & 45.23 & 75.87 & 69.19 & 52.63 & 77.51 & 63.80 \\
w/o cdl & 50.76 & 68.34 & 78.94 & 62.03 & 63.79 & 68.17 & 66.19 & 48.66 & 78.11 & 70.39 & 54.93 & 79.08 & 65.69 \\
 \hline
\end{tabular}
\end{adjustbox}
\caption{Classification accuracy (\%) for Partial Domain Adaptation Tasks on Office-Home dataset (backbone: Resnet-50).}

\vspace{-2mm}
\label{table_1}
\end{table*}

\begin{table*}[htbp!]
\centering
\begin{adjustbox}{width=0.55\textwidth}

 \begin{tabular}{c | c c c c c c | c} 
 \hline
 
Method & A $\rightarrow$ W & A $\rightarrow$ D &  W $\rightarrow$ A & W $\rightarrow$ D & D $\rightarrow$ A & D $\rightarrow$ W & Avg.\\
\hline

Resnet-50\cite{he2016deep} & 75.59 & 83.44 & 84.97 & 98.09 & 83.92 & 96.27 & 87.05\\
DAN\cite{long2015learning} & 59.32 & 61.78 & 67.64 & 90.45 & 74.95 & 73.90 & 71.34\\
DANN\cite{ganin2016domain} & 73.56 & 81.53 & 86.12 & 98.73 & 82.78 & 96.27 & 86.50\\
ADDA\cite{tzeng2017adversarial} & 75.67 & 83.41 & 84.25 & 99.85 & 83.62 & 95.38 & 87.03\\
PADA\cite{cao2018partial2} & 86.54 & 82.17 & 95.41 & \textbf{100.00} & 92.69 & \textbf{99.32} & 92.69\\
SSPDA\cite{cao2019learning} & 91.52 & 90.87 & 94.36 & 98.94 & 90.61 & 92.88 & 93.20\\
RTN\cite{long2016unsupervised} & 78.98 & 77.07 & 89.46 & 85.35 & 89.25 & 93.22 & 85.56\\
IWAN\cite{zhang2018importance} & 89.15 & 90.45 & 94.26 & 99.36 & \textbf{95.62} & \textbf{99.32} & 94.69\\
SAN\cite{cao2018partial} & 90.90 & \textbf{94.27} & 88.73 & 99.36 & 94.15 & \textbf{99.32} & 94.96\\

 \hline
Proposed model & \textbf{92.17} & 93.98 & \textbf{96.32} & \textbf{100.00} & 94.26 & 98.43 & \textbf{95.86}\\
w/o ast & 91.14 & 94.51 & 93.39 & 98.90 & 91.93 & 95.47 & 94.22\\
w/o adv & 83.39 & 81.19 & 89.04 & 98.63 & 87.98 & 97.31 & 89.59\\
w/o cdl & 85.71 & 84.03 & 90.12 & 98.26 & 88.97 & 96.13 & 90.53\\
\hline
\end{tabular}

\end{adjustbox}

\caption{Classification accuracy (\%) for Partial Domain Adaptation Tasks on Office-31 dataset (backbone: Resnet-50).}
\vspace{-8mm}
\label{table_2}
\end{table*}

$\hat{p}_t$ is utilized for the estimation of class importance through the confidence probabilities, $max\big(\hat{p}_t\big)$. It gives us an idea of the degree of likeliness with which the target sample $x_t$ is mapped to its closest cluster center. A low value indicates that the model is still confused about mapping $x_t$ to a category in $C_s$. Using unreliable target samples with low confidence values for class-importance weight estimation might thwart classification by misleading the model optimization task. Citing this, we devise a voting strategy to compute the class-importance weight vector $W$ (of size $|C_s|$), where a subset of the target samples (ones with high confidence predictions) are allowed to participate. This is mathematically represented as:

\begin{equation}
   D_t^{\mathcal{T}} = \{(x_t,\hat{y}^p_t) \hspace{1mm} | \hspace{1mm} x_t \in D_t, \hspace{1mm}max\big(\hat{p}_t\big) \geq \mathcal{T}\}
\end{equation}
\begin{equation}
\begin{gathered}
    W = \frac{W'}{max(W')}\\
    \text{where, } W' = \frac{1}{|D_t^{\mathcal{T}}|} \sum_{x_t \in D_t^{\mathcal{T}}} \hat{p}_t\\
\end{gathered}
\end{equation}

For a source sample $(x_s,y_s) \in D_s$, $w_{y_s}$ is represented as the corresponding class weight in the class-importance weight vector $W$. 

    The threshold parameter $\mathcal{T}$ is computed over the predicted outputs of the non-parametric classifier ${\Phi}(sim(\cdot))$ ($\Phi$ represents the softmax function) and measures the average probability of source domain samples belonging to the ground-truth class, i.e.:
    \begin{equation}
        \mathcal{T} = \frac{1}{n_s} \sum_{x_s \in D_s} max\big({\Phi}(sim(x_s))\big)
    \end{equation}

\subsubsection{\textbf{Entropy Minimization of Target Samples}}

In the early phases of a classification process, several side effects arise from adapting from one domain to another. They range from challenges with knowledge transfer caused by significant domain shifts to inducing uncertainty in the classifier. Entropy minimization on the predicted target samples forms a promising candidate to eliminate such adverse effects. In this work, we use an entropy minimization loss, which is described as: 
\begin{equation}
    L_{em} = - \frac{1}{n_t} \sum\limits_{x_t \in D_t} \sum\limits_{c_s \in C_s} \hat{y}^{c_s}_t log(\hat{y}^{c_s}_t)
\end{equation}

Where, $\hat{y}^{c_s}_t$ represents the probability of $x_t$ belonging to class $c_s$, as predicted by the classifier $class(\cdot)$. 

\subsubsection{\textbf{Overall Objective}}

To sum up, the overall objective function is modeled as follows:
\begin{equation}
\label{over_all_obj}
L = L_{class} + \alpha L_{adv} + \beta L_{var} + \gamma L_{em},
\end{equation}

with $\alpha, \beta$, and $\gamma$ as the trade-off hyper-parameters.

\section{\textbf{Experiments}}

In this section, we perform experiments on two benchmark datasets  (Office-Home\cite{venkateswara2017deep} and Office-31\cite{saenko2010adapting}) to evaluate the efficacy of the proposed framework. The experiments are conducted in a $pda$ setup over multiple tasks for each dataset. The following section reports the evaluation results, followed by an ablation analysis.

\subsection{\textbf{Datasets}} 

For the proposed technique's overall performance assessment, we use the standard datasets for domain adaptation, specifically \textit{Office-Home} and \textit{Office-31}.

The Office-31\cite{saenko2010adapting} dataset is relatively small and comprises 4652 images. These are grouped into 31 distinct categories, representing three domains: Amazon (A), DSLR (D), and Webcam (D). For evaluation purposes, we replicate the setup proposed by Cao et. al.  \cite{cao2018partial2}, where the target domain dataset contains images from 10 distinct classes. The assessment is conducted on 6 different permutations of source-target combinations, namely:  A$\rightarrow$W, A$\rightarrow$D, W$\rightarrow$A, W$\rightarrow$D, D$\rightarrow$A, and D$\rightarrow$W. 

The larger Office-Home dataset \cite{venkateswara2017deep}, utilized in this evaluation, is significantly more challenging. It contains a collection of 15,500 images grouped into 4 distinct domains: Artistic (Ar), Clip Art (Cl), Product (Pr), and Real-world (Rw). Following the PADA setup \cite{cao2018partial2}, we construct the source and target datasets with images from 65 and 25 different classes, respectively. 12 different permutations of source-target are used for evaluation purposes, namely Ar$\rightarrow$Cl, Ar$\rightarrow$Pr, Ar$\rightarrow$Rw, Cl$\rightarrow$Ar, Cl$\rightarrow$Pr, Cl$\rightarrow$Rw, Pr$\rightarrow$Ar, Pr$\rightarrow$Cl, Pr$\rightarrow$Rw, Rw$\rightarrow$Ar, Rw$\rightarrow$Cl, and Rw$\rightarrow$Pr.

\vspace{-0.5mm}

\subsection{\textbf{Implementation}}

The models in the evaluation are implemented using Pytorch on an Nvidia 3090-Ti GPU with 24 GB memory. We utilize Resnet-50 \cite{he2016deep}, pre-trained on the Imagenet dataset \cite{deng} and fine-tuned on the source data, as the backbone network for feature encoding. Before the fully-connected classification layers, we introduce a bottleneck layer of length 256. The discriminator consists of two fully-connected hidden layers of size 1024 with relu and 0.5 dropout probability, followed by the final layer of size 1 with sigmoid activation. The decoder comprises a fully-connected layer followed by a series of transpose convolution layers with batch normalization and leaky-relu, and sigmoid activations in the output layer. The network parameters are optimized using mini-batch SGD with a batch size of 36 and a momentum of 0.9 for 5000 epochs. The learning rates of the bottleneck, classification, decoding, and domain discriminator layers are 10 times that of the backbone, which is set to 1e-3 initially and adjusted as done in PADA \cite{cao2018partial2}.  We employ the same approach as DANN \cite{ganin2016domain} to introduce a gradient reversal layer for adversarial training and increase the $\alpha$ value from 0 to 1 as the number of iterations progresses. The trade-off parameters  $\beta$ and $\gamma$ are set to 0.8, 0.1, and 1, 0.1 for Office-31, and Office-home, respectively. We report the trainable classifier network outputs for target classification during model evaluation.

\subsection{\textbf{Comparison Models}}

The model is evaluated against state-of-the-art domain adaptation models specifically suitable for unsupervised closed-set and partial domain adaptation tasks, namely 
 Resnet-50 \cite{he2016deep}, Deep Adaptation Network (DAN) \cite{long2015learning}, Domain Adversarial Neural Network (DANN) \cite{ganin2016domain}, Adversarial Discriminative Domain Adaptation (ADDA) network \cite{tzeng2017adversarial}, Residual Transfer Networks (RTN) \cite{long2016unsupervised}, Importance Weighted Adversarial Nets (IWAN) \cite{zhang2018importance}, Selective Adversarial Network (SAN) \cite{cao2018partial}, Partial Adversarial Domain Adaptation(PADA) \cite{cao2018partial2} and class Subset Selection for Partial Domain Adaptation (SSPDA) \cite{cao2019learning}.

\section{\textbf{Results and Analysis}}

From the results summarized in Tables I and II, it is observed that the proposed method achieves comparable accuracy results to the state-of-the-art models addressing closed-set and partial domain adaptation on the presented tasks (achieving the highest accuracy in 8 out of 12 tasks  and in 3 out of 6 tasks on Office-Home and Office-31 datasets, respectively). It outperforms the compared models' overall average accuracy percentage on both datasets. 

Furthermore, we have also conducted an ablation analysis on the proposed framework by suppressing its three main components, one at a time:
\begin{itemize}

\item \textit{Proposed model without adaptive selection of target samples (ast)}: To evaluate its effectiveness, we limit the utilization of pseudo-labels for selecting highly confident target samples and computation of class-importance weights. Instead, we follow a strategy proposed by Cao et. al. \cite{cao2018partial} by aggregating the classifier prediction probabilities over all target samples to estimate $W$.

\item \textit{Proposed model without an adversarial loss (adv)}: To gauge the effectiveness of domain distribution alignment in latent space, we restrict the learning of domain-invariant latent representations by suppressing the adversarial objective (setting $\alpha$ of $L$ to 0 in eq. \ref{over_all_obj}).

\item \textit{Proposed model without class-distribution alignment (cdl)}: The proposed network utilizes the $L_{var}$ objective to regularize the latent space and achieve class distribution alignment. In this variant, we restrict the model from exploiting variational information for class distribution alignment by setting the $\beta$ value of $L$ to 0 (see eq. \ref{over_all_obj}).
\end{itemize}

The deterioration in the classification performance across all tasks (as witnessed in tables \ref{table_1} and \ref{table_2}) after suppressing these modules demonstrates their significance in the proposed framework and effectiveness in achieving better domain and class distribution alignments.

\section{\textbf{Conclusion}}

Improving class conditional distribution alignment forms a salient task in a $pda$ setup besides attaining domain invariance. Citing this, we couple an adversarial objective for domain alignment with a class distribution alignment strategy using variational information to regularize the latent space. Furthermore, we develop a robust technique for eliminating negative transfer and ensuring effective target supervision by adaptively selecting a subset of highly-confident target samples. The proposed model is tested under a range of $pda$ tasks against the state-of-the-art models addressing closed-set and partial domain adaptation problems for a comprehensive assessment. In addition, we performed an ablation analysis to verify the importance of the highlighted modules and establish their contribution to the suggested framework. The experimental findings demonstrate the suggested model's effectiveness over the compared models in the challenging tasks designed over two benchmark datasets.

\end{document}